\pdfoutput=1

\documentclass[conference]{IEEEtran}
\IEEEoverridecommandlockouts
\usepackage{amsmath,amssymb,amsfonts}
\usepackage{algorithmic}
\usepackage{graphicx}
\usepackage{textcomp}
\usepackage{xcolor}
\usepackage{booktabs}
\usepackage{url}
\usepackage{soul}
\usepackage[space]{cite}
\usepackage{pifont}
\newcommand{\cmark}{\ding{51}}%
\newcommand{\xmark}{}%
\usepackage{array}
\newcolumntype{P}[1]{>{\centering\arraybackslash}p{#1}}
\usepackage{cleveref}
\usepackage{listings}
\usepackage[frozencache]{minted}
\usepackage{csquotes}
\def\BibTeX{{\rm B\kern-.05em{\sc i\kern-.025em b}\kern-.08em
    T\kern-.1667em\lower.7ex\hbox{E}\kern-.125emX}}
    
\begin{document}

\title{Prompting and Fine-Tuning of Small LLMs for Length-Controllable Telephone Call Summarization\\
}

\author{
    David Thulke,\textsuperscript{\rm 1,2}
    Yingbo Gao,\textsuperscript{\rm 1}
    Rricha Jalota,\textsuperscript{\rm 1}
    Christian Dugast,\textsuperscript{\rm 1}
    Hermann Ney\textsuperscript{\rm 1,2}\\\,\\
    \textsuperscript{\rm 1}AppTek GmbH, Aachen \\
    \textsuperscript{\rm 2}Machine Learning and Human Language Technology Group, RWTH Aachen University \\
    \{dthulke, ygao, rjalota, cdugast, hney\}@apptek.com \\
}

\maketitle

\begin{abstract}
This paper explores the rapid development of a telephone call summarization system utilizing large language models (LLMs).
Our approach involves initial experiments with prompting existing LLMs to generate summaries of telephone conversations, followed by the creation of a tailored synthetic training dataset utilizing stronger frontier models.
We place special focus on the diversity of the generated data and on the ability to control the length of the generated summaries to meet various use-case specific requirements.
The effectiveness of our method is evaluated using two state-of-the-art LLM-as-a-judge-based evaluation techniques to ensure the quality and relevance of the summaries. Our results show that fine-tuned Llama-2-7B-based summarization model performs on-par with GPT-4 in terms of factual accuracy, completeness and conciseness.
Our findings demonstrate the potential for quickly bootstrapping a practical and efficient call summarization system.
\end{abstract}


\section{Introduction}

In many industries, particularly those involving customer service, healthcare, and finance, summarizing telephone conversations is a crucial task.
It is common practice for agents to summarize calls as part of call wrap procedures to ensure proper documentation and outline potential follow-ups.
This process can take around 10\% of the total call handling time, representing a significant operational overhead.

Besides reducing this overhead, automatic summarization of telephone calls offers several advantages: It can provide higher consistency and reduce the influence of individual agent biases. Additionally, automated summaries can still be reviewed and post-edited by agents if necessary, maintaining a balance between efficiency and accuracy.

Large language models (LLMs) have shown considerable promise in generating coherent and contextually relevant text.
Very large frontier LLMs like GPT-4~\cite{openai2023gpt4} and Llama 3.1 405B~\cite{dubey2024llama3herdmodels} have demonstrated strong zero-shot performance in various tasks, including summarization.
However, more efficient, smaller LLMs still struggle to match this level of performance. 
Task-specific training data may enable these smaller models to close the performance gap, offering a more resource-efficient alternative for practical applications.

This paper explores the rapid development of a telephone call summarization system utilizing LLMs.
Our approach involves initial experiments with prompting existing LLMs to generate summaries of telephone conversations.
These experiments provide insights into effective strategies and limitations, guiding the creation of a tailored training dataset using advanced frontier LLMs such as GPT-4.

A critical aspect of our research is the ability to control the length of the generated summaries to meet use-case specific requirements. We use Llama-2-7B~\cite{touvron2023llama2} as our base model. By fine-tuning the model and employing strategic prompting techniques, we aim to produce high-quality summaries that adhere to desired length constraints.

To evaluate the effectiveness of our method, we utilize state-of-the-art LLM-as-a-judge-based evaluation techniques. The results show that Llama-2-7B model, when fine-tuned on summarization-specific data is able to produce summaries that are on par with those generated by GPT-4 in terms of factual accuracy, completeness, and conciseness. In contrast, when the model is fine-tuned on task-agnostic, general data, its performance significantly deteriorates -- even lagging behind that of the Llama-2-Chat-7B model. 
This highlights the usefulness of training with task-specific data. Finally, our experiments on length-control reveal that training on uniform task-specific data can weaken the model's instruction-following ability. However, incorporating varying length-specific instructions into synthetic data generation can help restore this capability.
Our findings highlight the potential for quickly bootstrapping a practical and efficient call summarization system.

This paper is structured as follows:
\Cref{sec:data} introduces the telephone conversation corpus used in this work.
\Cref{sec:prompting} details the prompting techniques and synthetic data generation process.
\Cref{sec:finetuning} describes the models and approaches used for fine-tuning. \Cref{sec:evaluation} presents the evaluation methods and results, focusing on length adherence and utilizing LLM-as-a-judge evaluation.
\Cref{sec:related_work} reviews related work in the field of call summarization and LLMs.
Finally, \Cref{sec:conclusion} concludes the paper with a discussion of our findings and future research directions.

\section{Telephone Call Corpus}
\label{sec:data}

\begin{table}
    \centering
    \caption{Statistics of the Training, Validation, and Test splits of the simulated call-center calls}
    \label{table:data_splits}
    \begin{tabular}{lrrr}
        \toprule
        & \textbf{Train} & \textbf{Validation} & \textbf{Test} \\
        \midrule
        Number of Calls & 2,231 & 50 & 50 \\ 
        Average \# Turns per Call & 164.3 & 190.5 & 131.0 \\ 
        Average \# Words per Turn & 17.7 & 15.6 & 18.3 \\ 
        \# of unique words & 44,506 & 6,918 & 6,268 \\
        \bottomrule
    \end{tabular}
\end{table}

\begin{table}
    \centering
    \caption{Distribution of accents in the Training, Validation, and Test splits of the simulated call-center calls}
    \label{table:data_accents}
    \begin{tabular}{lrrr}
        \toprule
        & \textbf{Train} & \textbf{Validation} & \textbf{Test} \\
        \midrule
        African American & 504 & 12 & 14 \\
        Chinese & 140 & 5 & 0 \\
        Hispanic & 1,018 & 16 & 25 \\
        Southern & 569 & 17 & 11 \\
        \bottomrule
    \end{tabular}
\end{table}

For this work, we utilized a corpus of 2,331 simulated telephone recordings. We did not use real customer conversations to avoid any privacy concerns.
The simulated corpus was originally created to augment AppTek's automatic speech recognition training data, focusing specifically on US English accented speech, including accents from African American, Hispanic, Chinese, and Southern speakers.

The recordings in this corpus were performed by 479 speakers from AppTek's data workforce representing a range of accents.
Each pair of speakers was provided with a specific topic to discuss during their conversation.
Topics included for example travel, insurance and movies, ensuring a wide variety of conversational content.

\Cref{table:data_splits} contains detailed statistics on the training, validation, and test data used in our experiments.
Additionally, \Cref{table:data_accents} shows the distribution of accents within the data.
By using this diverse and controlled dataset, we were able to systematically explore the performance of our telephone call summarization system without compromising privacy or relying on actual customer data.

\section{Synthetic Data Generation and Prompting}
\label{sec:prompting}

Due to the high costs and effort involved in human data annotation, we opted to generate synthetic summaries of call transcripts using GPT-4~\cite{openai2023gpt4}, which is a strong external model capable of producing high-quality summaries. This approach allowed us to automatically generate a large volume of summaries for model supervision while minimizing costs. 

To generate high-quality summaries, we first preprocess the call transcripts by extracting speaker tags from the raw data and incorporating them into the transcript. The inclusion of speaker information aims to provide additional context that the model can utilize when generating summaries. While our system is entirely text-based, this step also considers potential future applications, such as multimodal models that process audio inputs from call recordings to produce text summaries, where speaker information could be valuable. We anonymize the speaker tags, labeling them as \texttt{speaker 0} and \texttt{speaker 1} when two speaker channels are present. Although more detailed speaker information could be included, it falls outside the scope of this work. 

In addition to including speaker tags, we also account for longer context lengths. Given that GPT-4 supports a larger context window (8k) compared to our Llama2~\cite{touvron2023llama2} base model (4k), generating synthetic data with a context size larger than our model’s capacity is inefficient. Therefore, when necessary, we truncate the left side of the context to ensure that the combined call transcript and summarization prompt fit within our model's context window. While this may impact summarization quality, it is a necessary compromise  we have to make. 

Prompting is an important part of the LLM pipeline as it aligns the model behavior with the user purpose~\cite{chen2024unleashingpotentialpromptengineering}. Due to our above-mentioned context pruning method and findings from our preliminary experiments—where prompts positioned closer to the current content produced better results compared to those positioned further away (e.g., “Summarize the call transcript above” at the end yielded better outcomes than “Summarize the call transcript below” at the beginning)—we adopt a transcript-first-and-prompt-last format for prompting.

We define three categories for summarization-specific prompts, namely \enquote{default}, \enquote{general} and \enquote{length-oriented}. Detailed prompts can be found in Table \ref{summarization_prompts}. The \enquote{default} category includes a straightforward command that instructs the model to summarize the call transcript. This serves as the baseline prompt.
For the \enquote{general} category, we used GPT-4 to generate a range of summarization prompts focusing on various aspects such as content, sentiment, next steps, and so on. We then reviewed these prompts manually and selected those that met our quality criteria.

Finally, for the \enquote{length-oriented} category, we used six prompt variants, each imposing specific constraints in sentence count, word count or paragraph count to control the summary length.
The objective of this category was to train the model to adhere to length-specific instructions. We anticipated that exposure to such length-specific prompts during training would enable the model to follow these instructions, thereby providing a 'soft' mechanism for length control during testing.

This way, in total, we considered 20 summarization-specific prompts, and for each transcript during training, we randomly sampled five variants from this pool of prompts.

\begin{table*}
\label{summarization_prompts}
\caption{List of summarization-specific prompts in the Default, General and Length-specific category used in this work}
\centering
\begin{tabular}{l|l}
\toprule
\textbf{Category}
& \textbf{Prompts} \\ \midrule
Default 
& \hphantom{1}1.\quad Summarize the call transcript above. \\ \midrule
General 
& \begin{tabular}[c]{@{}l@{}}
    \hphantom{1}1.\quad From the call transcript above, extract and summarize important points about 1. call intent, 2. next steps, 3. outcome.\\
    \hphantom{1}2.\quad Summarize the key issues and resolutions discussed in this call center transcript.\\
    \hphantom{1}3.\quad Provide a brief summary of the customer’s issue and the call center agent’s response from the transcript.\\
    \hphantom{1}4.\quad Extract and summarize the main points of discussion, including any action items, from this call transcript.\\
    \hphantom{1}5.\quad Generate a brief overview of the call, highlighting any commitments or follow-ups mentioned in the transcript.\\
    \hphantom{1}6.\quad Analyze this call transcript and summarize the outcome of the customer service interaction.\\ 
    \hphantom{1}7.\quad Summarize the steps taken by the agent to address the customer's issue in this call transcript.\\ 
    \hphantom{1}8.\quad Briefly summarize the key facts of the customer’s inquiry and the agent’s assistance from this transcript.\\ 
    \hphantom{1}9.\quad From the transcript, summarize the customer's feedback and how the call center agent handled it.\\ 
    10.\quad Summarize the customer’s issue and the steps discussed in the call transcript.\\ 
    11.\quad From the transcript, create a summary of any technical issues reported and the solutions provided.\\ 
    12.\quad Create a summary of the call transcript, focusing on the customer satisfaction level by the end.\\ 
    13.\quad Summarize the emotional tone of the customer and the empathy expressed by the agent in the call transcript.
\end{tabular} \\ \midrule
Length 
& \begin{tabular}[c]{@{}l@{}}
    \hphantom{1}1.\quad Summarize the call transcript above in one sentence.\\ 
    \hphantom{1}2.\quad Summarize the call transcript above in two sentences.\\ 
    \hphantom{1}3.\quad Summarize the call transcript above in 50 words.\\ 
    \hphantom{1}4.\quad Summarize the call transcript above in 100 words.\\ 
    \hphantom{1}5.\quad Summarize the call transcript above in one paragraph.\\ 
    \hphantom{1}6.\quad Summarize the call transcript above in two paragraphs.
\end{tabular} \\ \bottomrule
\end{tabular}
\end{table*}

Next, we selected a simple and straightforward system prompt for telephone call summarization: \enquote{You are good at summarizing call transcripts.}
It is arguable if such a system prompt is even needed for a task-specific model, especially when alternative system prompts are foreign to the model.
However, considering that our models also see general instruction data during training (described in Section~\ref{sec:finetuning}), i.e. other instructions that are not summarization-specific, we nonetheless include this system prompt to better align the model to the summarization objective.

\section{Fine-Tuning}
\label{sec:finetuning}

In this section, we discuss the fine-tuning process that we employed to enhance the model's summarization performance.
As we want to preserve the general-domain instruction following capabilities of the model, we also train our model on non-summarization-specific instruction fine-tuning data.
We start this section by describing this data followed by an overview of our training setup.

\subsection{General-domain Instruction Fine-Tuning Data}

Besides the summarization-specific instruction fine-tuning (IFT) data, we extend the training data with general-domain data to improve the instruction following capabilities and robustness of the model. We refer to this data as \emph{Our} IFT data and use the same data mixture described in our previous work \cite{thulke2024climategpt}.
Specifically, we include the following subsets:
\begin{itemize}
    \item an internal high-quality set of 700 prompt-completion pairs originally collected by AppTek
    \item \emph{Databricks Dolly} \cite{DatabricksBlog2023DollyV2} the first openly available human-generated IFT dataset with a permissive license consisting of 15,001 prompt and completion pairs across 7 task categories
    \item the English subset of \emph{OpenAssistant Conversations 1} (OASST-1) \cite{koepf2023openassistant} taking only the best-rated conversations, resulting in 3,783 conversations.
    \item 3,282 question-and-answer pairs from \emph{StackExchange}
    \item a subset of 45,000 samples from \emph{FLAN v2 and CoT} \cite{wang2023how}
    \item \emph{Llama-2 Safety} consisting of 939 refusals to prompts in the Do-Not-Answer dataset \cite{wang-etal-2024-answer} synthetically generated using Llama-2-Chat-70B \cite{touvron2023llama}
\end{itemize}
In the following we refer to all Llama-2-7B models that are fine-tuned on this data as \emph{Llama-2-7B-Our}.

\subsection{Summarization-specific Instruction Fine-Tuning Data}
Using the three categories of summarization-specific prompts outlined in Table~\ref{summarization_prompts} and following the methodology described in Section~\ref{sec:prompting}, we generate three distinct types of summarization-specific IFT data using GPT-4: 
\emph{Default}, \emph{General}, and \emph{Length}-specific Summarization IFT, which contain 580, 7,248, and 3,327 instances, respectively.
We then fine-tune Llama-2-7B~\cite{touvron2023llama2} using various combinations of this data, resulting in seven different summarization-specific IFT model variants, as shown in Table~\ref{tab:llm-as-a-judge}.

\subsection{Training Setup}

We use a fork of NVIDIA's Megatron-LM \cite{nvidia_megatron} by the EPFL LLM Team \cite{epfmgtrn, chen2023meditron70b} for IFT training.
We use a cosine learning rate schedule with a peak learning rate of $10^{-5}$ and a warm-up of 100 steps.
The batch size is set to 64 and we use the full sequence length of 4096 tokens.
For regularization, we use weight decay of $10^{-2}$ and dropout as used for LIMA \cite{zhou2023lima}.
Since the exact summarization instruction data seen by each model variant differs, the number of training steps also differ.
That said, because our general-domain instruction data (in comparison to summarization-specific instruction data) still makes up the majority of the total instruction data, this difference is not significant among the models, ranging from 1814 training steps to 2337 training steps.
All models are trained on 4xA100 80GB GPUs utilizing pipeline parallelism and using full weight fine-tuning.

We use Chat Markup Language\footnote{\url{https://github.com/openai/openai-python/blob/release-v0.28.0/chatml.md}}\footnote{\url{https://learn.microsoft.com/en-us/azure/ai-services/openai/how-to/chat-markup-language}} as prompt template to format the IFT data, which results in the following format:
\begin{verbatim}
<|im_start|>system
[system_prompt]<|im_end|>
<|im_start|>user
[call_transcript]
[summarization_prompt]<|im_end|>
<|im_start|>assistant
[call_summary]<|im_end|>
\end{verbatim}
The fields in square brackets (including the brackets themselves) are replaced by the actual data. During training, the call summary is replaced by the summary produced by GPT-4.
During testing, the model is provided with the prompt until \texttt{assistant}
and the model generates both the summary and the special end symbol.

\section{Evaluation}
\label{sec:evaluation}

For evaluation, we focus on two aspects.
First, the general quality of the generated summaries is evaluated using LLM-as-a-Judge approaches.
Second, we analyze the effect of our approaches on the length adherence of the resulting models.

As baselines, we consider GPT-4~\cite{openai2023gpt4} and Llama-2-Chat-7B~\cite{touvron2023llama}, which was instruction fine-tuned on general domain data as well as further tuned using reinforcement learning from human feedback. We perform greedy decoding and set the maximum completion length to 256. 
If not stated otherwise all models are prompted with the \emph{default} summarization prompt described in \Cref{summarization_prompts}.

\subsection{LLM-as-a-Judge}

We evaluate our models using LLM-as-a-judge based evaluation methods, specifically, Prometheus-Eval~\cite{kim2024prometheus} and FineSurE~\cite{song-etal-2024-finesure}. These approaches provide a framework for evaluating models against various task-relevant criteria. For summaries to be useful, it is important that they contain a gist of all the key facts discussed in the call, without any misleading information. To ensure this, we perform a multi-dimensional evaluation. 

For evaluation with Prometheus-Eval~\cite{kim2024prometheus}, we utilize two predefined rubrics to determine if the summaries are non-misleading (\texttt{HONESTY}) and factually accurate (\texttt{FACTUAL\_VALIDITY}). Additionally, we introduce a custom rubric to assess if a summary includes all the main points discussed in the call (\texttt{COMPLETENESS}) (see Figure~\ref{fig:completeness}). The summaries are evaluated using the Prometheus-8x7B model~\cite{kim2024prometheus}, which assigns a Likert-scale score from 1 to 5 for each rubric per summary. 

\begin{figure}
    \begin{minted}[frame=single,
                   breaklines,
                   breaksymbolleft=\ ,
                   framesep=3mm,
                   linenos=true,
                   numbers=none,
                   fontsize=\scriptsize,
                   tabsize=2]{js}
{
  "criteria": "Does the model's response cover all the key points discussed in the call with sufficient context to understand them?",
  "score1_description": "The model's response fails to cover the key points and lacks sufficient context.",
  "score2_description": "The model's response covers a few key points but lacks sufficient context.",
  "score3_description": "The model's response covers some key points with some context but is incomplete.",
  "score4_description": "The model's response covers most key points with sufficient context.",
  "score5_description": "The model's response fully covers all key points with complete and clear context."
}
    \end{minted}
\caption{Definition of the completeness rubric in Prometheus-Eval} 
\label{fig:completeness}
\end{figure}

FineSurE~\cite{song-etal-2024-finesure}, unlike Prometheus-Eval, is specifically designed for fine-grained summarization evaluation. It assesses summaries from three perspectives: Faithfulness, Completeness, and Conciseness. \texttt{FAITHFULNESS} is evaluated by determining the proportion of sentences in the summary that are factually correct, ensuring the summary is free from factual errors. \texttt{COMPLETENESS} is measured as the ratio of the number of key facts from the transcript that are effectively covered in the summary to the total number of key facts extracted from the summary, ensuring the summary is thorough. \texttt{CONCISENESS} is determined by the ratio of the number of sentences that contain relevant key facts to the total number of sentences in the summary, thereby evaluating the brevity of the summary. We report the percentage scores for each metric. Key fact extraction and evaluation of the summaries is done using GPT-4o (gpt-4o-2024-05-13). The default prompt for key fact extraction in FineSurE requires a reference summary. However, since we do not have reference summaries in our test set, we extract key facts directly from the call transcript, using the prompt template shown in Figure~\ref{fig:keyfact}.

\begin{figure}
    \begin{minted}[frame=single,
                   breaklines,
                   breaksymbolleft={},
                   framesep=3mm,
                   linenos=true,
                   xleftmargin=2pt,
                   numbers=none,
                   fontsize=\scriptsize,
                   tabsize=2]{text}
You will be provided with a transcript that might contain diarization errors. Your task is to extract a set of "key facts" from the transcript that adequately summarize it. Ensure that the key points summarize the main ideas and important details discussed. A "key fact" is a single sentence written as briefly and clearly as possible, encompassing the essence of the conversation, highlighting any significant facts, decisions, or conclusions mentioned. Aim for clarity and conciseness, avoiding minor details or tangential topics.

Instruction:
First, read the transcript carefully.
Second, decompose the transcript into (at most 16) key facts.

Provide your answer in JSON format. The answer should be a dictionary with the key "key_facts" containing the key tacts as a list:
{"key_facts" ["first key fact", "second key fact", "third key fact"]}

Transcript:
{transcript}
    \end{minted}
    \caption{Prompt for Key Fact Extraction from Call Transcript}
    \label{fig:keyfact}
\end{figure}

\begin{table*}
    \centering
    \caption{Evaluation with LLM-as-a-judge approaches: FineSurE and Prometheus-Eval. \\Bold numbers highlight the best overall score per metric, while the second-best scores are underlined.}
    \label{tab:llm-as-a-judge}
    \centering
    \begin{tabular}{lccc|rrr|rrr|r}
    \toprule
     & \multicolumn{3}{c}{\textbf{Summarization IFT}} & \multicolumn{3}{|c}{\textbf{FineSurE}} & \multicolumn{3}{|c}{\textbf{Prometheus 8x7B}} & \multicolumn{1}{|c}{} \\
    \cmidrule(lr){2-4} \cmidrule(lr){5-7} \cmidrule(lr){8-10}
    \textbf{Model}      & \scriptsize \textbf{Default} &\scriptsize \textbf{General} &\scriptsize \textbf{Length}
          & {\scriptsize\textbf{Faithfulness}} & \scriptsize\textbf{Completeness} &\scriptsize \textbf{Conciseness}  & {\scriptsize\textbf{Factual Validity}} & \scriptsize\textbf{Honesty} & \scriptsize\textbf{Completeness} & \textbf{Avg.} \\ \midrule
    GPT-4  &&& & {\textbf{88.90}} & 30.50   & {\textbf{70.70}} & \textbf{4.22}   & \ul{4.56}  & 3.80  & \textbf{33.78} \\
    Llama-2-Chat-7B &&& & 63.30   & 24.50  & 50.90  & 3.84 & 4.22  & 3.60 & 25.06 \\ \midrule
    Llama-2-7B-Our 
    & \xmark & \xmark & \xmark & 60.00 & 13.10 & 41.60   & 2.60  & 3.06  & 2.46& 20.47 \\
    & \cmark & \xmark & \xmark  & 78.30 & 36.20  & 64.30  & 4.04  & \ul{4.56}   & 3.72  & 31.85 \\
    & \xmark & \cmark & \xmark  & \ul{86.00}  & 32.30     & 61.30  & 4.08 & 4.34   & 3.74 & 31.96 \\
    & \xmark & \xmark & \cmark & 81.70 & 34.80  & \ul{70.30} & \ul{4.12}  & \textbf{4.64} & 3.76 & 33.22 \\
    & \cmark & \cmark & \xmark & 81.20  & 36.00  & 63.10 &  4.06  & 4.40  & \textbf{3.88}  & 32.11 \\
    & \cmark & \xmark & \cmark & 81.40   & \textbf{37.70}  & 67.60  & {3.98} & 4.46  & 3.76& 33.15 \\
    & \xmark & \cmark & \cmark & 80.40  & 35.20    & 66.90    & {4.08}  & 4.54   & \ul{3.82} & 32.49 \\
    & \cmark & \cmark & \cmark & {84.40} & \ul{35.70} & 67.00  & \ul{4.12} & 4.46    & 3.74  & \ul{33.24}\\ 
    \bottomrule
    \end{tabular}%
    \end{table*}

 Table~\ref{tab:llm-as-a-judge} presents the results with average scores per metric. Bold numbers highlight the best overall score per metric, while the second-best scores are underlined.
We first note that the Llama-2-7B-Our model trained only on the general-domain IFT data slightly lacks behind Llama-2-Chat-7B across all metrics.
This can be explained by the larger amount of IFT data and additional RLHF fine-tuning that Llama-2-Chat-7B received.
Including summarization-specific IFT data significantly improves the performance of the model across all metrics with results that are on-par with GPT-4.
While our models still slightly lagging behind GPT-4 in faithfulness and factual validity but surpassing it in completeness and honesty.
Notably, all variants of our summarization IFT models significantly outperform Llama-2-Chat-7B across all metrics from both FineSurE and Prometheus-Eval.
Taking an average over all metrics, our best-performing model is the one trained on all three summarization-specific IFT categories, only slightly behind GPT-4.

Looking at the Completeness that is the only dimension in common of both evaluation frameworks, we observe some differences in the ranking of the best performing systems.
Analyzing this in more detailed reveals that the FineSurE completeness results negatively correlates with the length of the telephone calls.
The longer the call, the higher the probability that key facts from the call are not covered in the summary.
As Prometheus-Eval performs the evaluation in a single step, it is less affected by the length of the call, and we do not observe any correlation to the call length.

Since Prometheus-8x7B was not trained on our manually defined \texttt{COMPLETENESS} rubric, we conducted an additional evaluation of the summaries generated by the baseline models and a selected IFT variant using GPT-4 within the Prometheus-Eval framework.
Among the summarization IFT model variants, although there is no clear winner, we choose Llama-2-7B-Our trained on all three categories of summarization-specific IFT data for further evaluation with GPT-4.
This choice was made based on its relatively superior performance across all metrics.
Results in \Cref{tab:gpt4-eval} demonstrate that the summaries generated by our selected IFT model are comparable in quality to those produced by GPT-4.
This further indicates that task-specific data, when used with smaller and more resource-efficient models, can help bridge the performance gap.
We present some example summaries from the three models in \Cref{tab:summaries}.

\begin{table}
    \centering
    \caption{Evaluation with GPT-4 as a judge. Llama-2-7B-Our Summ.\ is trained on all three types of summarization IFT data.}
    \label{tab:gpt4-eval}
    {%
    \begin{tabular}{l|rrr}
        \toprule
        & \multicolumn{3}{c}{\textbf{GPT-4}} \\
        \cmidrule(lr){2-4}
        \textbf{Model} & \scriptsize \textbf{Factual Validity} & \scriptsize\textbf{Honesty} & \scriptsize\textbf{Completeness} \\        \midrule
        GPT-4 & \textbf{4.80} & 4.78 & 4.52 \\
        Llama-2-Chat-7B & 3.98 & 4.48 & 4.14 \\ \midrule
        Llama-2-7B-Our Summ. & \textbf{4.80} & \textbf{4.90} & \textbf{4.68} \\
        \bottomrule
    \end{tabular}
    }
\end{table}

\subsection{Length Adherence}

\begin{table}
    \centering
    \caption{Length Adherence: percentage of summaries generated by Llama-2-7B-Our models that adhere to the length constraint given in the prompt.}
    \label{tab:length}
    \begin{tabular}{ccc|rr}
        \toprule
        \multicolumn{3}{c}{\textbf{Summarization IFT}}  & &\\
        \cmidrule(lr){1-3}
        \scriptsize \textbf{Default} &\scriptsize \textbf{General} &\scriptsize \textbf{Length} & 50 words & 100 words \\
        \midrule
        \xmark & \xmark & \xmark &  44\% & \textbf{68\%} \\
        \cmark & \xmark & \xmark &  0\% & 28\%   \\
        \xmark & \cmark & \xmark &   2\% & 20\% \\
        \xmark & \xmark & \cmark &   44\% & 48\%  \\
        \cmark & \cmark & \xmark &   0\% & 22\%  \\
        \cmark & \xmark & \cmark &  42\% & 58\% \\
        \xmark & \cmark & \cmark &   \ul{50\%} & 54\% \\
        \cmark & \cmark & \cmark &  \textbf{56\%} &  \ul{60\%}  \\
        \bottomrule
    \end{tabular}
\end{table}

We further assess the effectiveness of length-oriented summarization IFT by testing the IFT variants with prompts that include specific length constraints and report the percentage of summaries that adhered to these constraints.
The prompts used for this evaluation are: \textit{Summarize the call transcript above in 50 words} and \textit{Summarize the call transcript above in 100 words} which were both part of the length-specific summarization IFT data. 

The results are shown in \Cref{tab:length}.
We first observe that the model trained on general-domain IFT data only (\textit{Llama-2-7B-Our}) adheres to the length constraints.
When incorporating summarization-specific IFT data without length constraints (Default and General), the model's adherence to the length constraints significantly degrades.
We assume this is caused by the model overfitting to summarization lengths that occur in this data.
However, incorporating length-specific instructions into the summarization-specific data restores this capability.
Finally, we observe that even the best performing models only achieve adherence rates of around 60\%, indicating that there is still room for improvement in this area.
Possible solutions could include filtering the synthetically generated data or incorporating even more and diverse length-specific data.

\begin{table*}
    \centering
    \caption{Example Summaries}
    \label{tab:summaries}
    \begin{tabular}{p{0.31\linewidth}|p{0.31\linewidth}|p{0.31\linewidth}}
    \toprule
      \multicolumn{1}{c|}{\textbf{GPT-4}} & \multicolumn{1}{c|}{\textbf{Llama-2-Chat-7B}} & \multicolumn{1}{c}{\textbf{Llama-2-7B-Our Summ.}} \\ \midrule \scriptsize
      Speaker 1, a representative of an airlines, confirms a first-class seat upgrade for Speaker 2, who is traveling alone on business to Alaska. Speaker 2 shares that they will be representing a startup ice cream company at a convention in Alaska. Speaker 1 verifies that the upgrade includes inflight meals, or an alternative of snacks, early check-in, and the journey is from 22nd to 28th, while using only a portion of Speaker 2's reward points. Speaker 2 appreciates Speaker 1's help. The call ends with mutual courteous exchanges.
  &   \scriptsize A customer named Jane Leveque called $<$initial$>$KNW$<$/initial$>$ Airlines to upgrade her seat from economy to first class for her upcoming flight from Boston to Anchorage, Alaska. The customer service representative, speaker 1, was able to upgrade Jane's seat using 500,000 of her 1.5 million reward points. Jane was also able to choose her in-flight meal in advance and will have early check-in for her flight. The representative confirmed the details of Jane's flight and thanked her for calling $<$initial$>$KNW$<$/initial$>$ Airlines.
 & \scriptsize Speaker 1, an airline representative, is assisting Speaker 2, Jane Leveque, with a seat upgrade for her upcoming flight from Boston to Alaska. Jane has a significant amount of reward points, 1.5 million, which allows her to upgrade to first class without using all of her points. The representative explains the benefits of the first-class ticket, including two baggage check-ins, priority TSA line, free in-flight meal, and free beverages. Jane plans to represent a new ice cream company at a convention in Alaska and is excited about the opportunity. The representative confirms that Jane can check in for her flight anytime and also offers snacks in addition to the meal. Jane appreciates the representative's help and confirms that she has no more questions. The representative thanks Jane for choosing their airline and wishes her a great flight.
 \\ \midrule
 \scriptsize In the conversation, Speaker 1 and Speaker 2 discuss their thoughts about certain movies. Speaker 2 starts by expressing dissatisfaction about "La Llorona", critiquing its American setting and the movie's excessive use of screaming. She suggests the film would have been better if set in Mexico and lamented that it wasn't as expected. Adding to the conversation, Speaker 1 expresses disappointment about a few films, mentioning "Soul Plane" and "Snakes on a Plane" as two they didn't enjoy. Speaker 1 criticizes specific plot points as absurd and the overall quality poor. Toward the end of the conversation, the idea of a movie night with friends in two weeks is proposed by Speaker 2. Speaker 1 agrees to the plan and suggests Speaker 2 make a call once the date is set. The call ends on friendly terms.
 & \scriptsize     Two friends, Speaker 1 and Speaker 2, had a conversation about their favorite horror movies and shared their opinions on different films. They discussed various movies, including "The Others," "Strangers," "Poltergeist," and "The Never Ending Story." They also shared their thoughts on love stories and mentioned a few movies they didn't enjoy, such as "La Llorona" and "Soul Plane." The conversation was lighthearted and friendly, with the two friends expressing their enthusiasm for movie nights and sharing their love for cinema.
 &  \scriptsize Speakers 1 and 2 discuss their favorite movies, including horror, thriller, and romantic genres. Speaker 1 shares their love for the movie "Tuck Everlasting," while Speaker 2 expresses disappointment in the movie "La Llorona." They also discuss the movies "The Never Ending Story," "The Exorcist," and "Drag Me to Hell." Towards the end of the conversation, they plan a movie night with friends and agree to coordinate their schedules in two weeks.
 \\ \bottomrule
    \end{tabular}
\end{table*}

\section{Related Work}
\label{sec:related_work}

While there is some previous work studying both extractive \cite{tamura-etal-2011-extractive} as well as abstractive \cite{favre-etal-2015-call} telephone call summarization, most recent work using LLMs focuses one news summarization \cite{goyal2022news,zhang-etal-2024-benchmarking}.
Most related to our work is recent efforts on using LLMs for written dialog summarization \cite{ramprasad-etal-2024-analyzing,chen-etal-2021-dialogsum}.
Asi et al.\ utilize commercial API models for telephone conversation \cite{asi-etal-2022-end} while in our approach we distill the summarization capabilities of a frontier LLM to a small LLM.
Concurrently to our work, Mullick et al.\ explore the use of synthetic data for domain-specific document summarization \cite{mullick-etal-2024-persona}.

Length control is a common requirement for abstractive summarization systems.
While LLMs allow to define arbitrary constraints in natural language, previous systems often relied on special control tokens \cite{fan-etal-2018-controllable}.
In addition to supervised fine-tuning, previous work also utilized reinforcement learning \cite{jie-etal-2024-prompt} or preference optimization \cite{yuan2024followinglengthconstraintsinstructions} to enable length control in text generation models.

\section{Conclusion}
\label{sec:conclusion}

In this paper, we explored the rapid development of a telephone call summarization system utilizing LLMs, with a particular emphasis on achieving high performance by utilizing small models.
Our experiments demonstrate that it is possible to significantly close the gap between small use-case-specific LLMs and frontier models like GPT-4, even in the absence of use-case-specific data.
By generating a tailored synthetic training dataset, we are able to fine-tune a 7B parameter model to achieve summarization capabilities approaching or even slightly exceeding the performance of GPT-4 in key metrics.
Our results show that the diversity of the prompts in the synthetically generated data is crucial to improving performance — even if just a single prompt is used for evaluation.
Further, our analysis and experiments on length control show that training on homogeneous task-specific data can deteriorate instruction following capabilities.
We show that, at least for length control, this specific capability can be recovered by augmenting the synthetic data generation.

While our results are promising, they represent an initial step in the broader development of a robust call summarization system.
Future research should focus on addressing several important areas that were not fully explored in this study. 
These include 
the impact of speech recognition errors,
the impact of text normalization (e.g. conversion of spoken to written numbers) and
the impact of personal identifiable information that is often redacted in contact center transcripts.
Additionally, previous work has shown that pre-training or continued pre-training on in-domain data (i.e. call transcripts in case) can improve the downstream performance of LLMs \cite{Wu2023bloombergGPT, chen2023meditron70b, thulke2024climategpt}.
Further, while our synthetic dataset was effective for fine-tuning, real-world data from the specific domain of application is expected to provide additional benefits, leading to even higher levels of performance and reliability.
Finally, while LLM-as-a-judge approaches provide a useful evaluation framework, human evaluation is still necessary to assess the real-world utility and to make deployment decisions.

\clearpage 
\bibliographystyle{IEEEtran}
\bibliography{anthology,custom}

\begin{thebibliography}{10}
\providecommand{\url}[1]{#1}
\csname url@samestyle\endcsname
\providecommand{\newblock}{\relax}
\providecommand{\bibinfo}[2]{#2}
\providecommand{\BIBentrySTDinterwordspacing}{\spaceskip=0pt\relax}
\providecommand{\BIBentryALTinterwordstretchfactor}{4}
\providecommand{\BIBentryALTinterwordspacing}{\spaceskip=\fontdimen2\font plus
\BIBentryALTinterwordstretchfactor\fontdimen3\font minus \fontdimen4\font\relax}
\providecommand{\BIBforeignlanguage}[2]{{%
\expandafter\ifx\csname l@#1\endcsname\relax
\typeout{** WARNING: IEEEtran.bst: No hyphenation pattern has been}%
\typeout{** loaded for the language `#1'. Using the pattern for}%
\typeout{** the default language instead.}%
\else
\language=\csname l@#1\endcsname
\fi
#2}}
\providecommand{\BIBdecl}{\relax}
\BIBdecl

\bibitem{openai2023gpt4}
OpenAI, ``Gpt-4 technical report,'' 2023.

\bibitem{dubey2024llama3herdmodels}
\BIBentryALTinterwordspacing
Meta, ``The llama 3 herd of models,'' 2024. [Online]. Available: \url{https://arxiv.org/abs/2407.21783}
\BIBentrySTDinterwordspacing

\bibitem{touvron2023llama2}
\BIBentryALTinterwordspacing
H.~Touvron, L.~Martin, K.~Stone, P.~Albert, A.~Almahairi, Y.~Babaei, N.~Bashlykov, S.~Batra, P.~Bhargava, S.~Bhosale \emph{et~al.}, ``Llama 2: Open foundation and fine-tuned chat models,'' \emph{ArXiv preprint}, vol. abs/2307.09288, 2023. [Online]. Available: \url{https://arxiv.org/abs/2307.09288}
\BIBentrySTDinterwordspacing

\bibitem{chen2024unleashingpotentialpromptengineering}
\BIBentryALTinterwordspacing
B.~Chen, Z.~Zhang, N.~Langrené, and S.~Zhu, ``Unleashing the potential of prompt engineering in large language models: a comprehensive review,'' 2024. [Online]. Available: \url{https://arxiv.org/abs/2310.14735}
\BIBentrySTDinterwordspacing

\bibitem{thulke2024climategpt}
\BIBentryALTinterwordspacing
D.~Thulke, Y.~Gao, P.~Pelser, R.~Brune, R.~Jalota, F.~Fok, M.~Ramos, I.~van Wyk, A.~Nasir, H.~Goldstein, T.~Tragemann, K.~Nguyen, A.~Fowler, A.~Stanco, J.~Gabriel, J.~Taylor, D.~Moro, E.~Tsymbalov, J.~de~Waal, E.~Matusov, M.~Yaghi, M.~Shihadah, H.~Ney, C.~Dugast, J.~Dotan, and D.~Erasmus, ``Climategpt: Towards ai synthesizing interdisciplinary research on climate change,'' 2024. [Online]. Available: \url{https://arxiv.org/abs/2401.09646}
\BIBentrySTDinterwordspacing

\bibitem{DatabricksBlog2023DollyV2}
\BIBentryALTinterwordspacing
M.~Conover, M.~Hayes, A.~Mathur, J.~Xie, J.~Wan, S.~Shah, A.~Ghodsi, P.~Wendell, M.~Zaharia, and R.~Xin. (2023) Free dolly: Introducing the world's first truly open instruction-tuned llm. [Online]. Available: \url{https://www.databricks.com/blog/2023/04/12/dolly-first-open-commercially-viable-instruction-tuned-llm}
\BIBentrySTDinterwordspacing

\bibitem{koepf2023openassistant}
\BIBentryALTinterwordspacing
A.~K{\"o}pf, Y.~Kilcher, D.~von R{\"u}tte, S.~Anagnostidis, Z.~R. Tam, K.~Stevens, A.~Barhoum, D.~M. Nguyen, O.~Stanley, R.~Nagyfi, S.~ES, S.~Suri, D.~A. Glushkov, A.~V. Dantuluri, A.~Maguire, C.~Schuhmann, H.~Nguyen, and A.~J. Mattick, ``Openassistant conversations - democratizing large language model alignment,'' in \emph{Thirty-seventh Conference on Neural Information Processing Systems Datasets and Benchmarks Track}, 2023. [Online]. Available: \url{https://openreview.net/forum?id=VSJotgbPHF}
\BIBentrySTDinterwordspacing

\bibitem{wang2023how}
\BIBentryALTinterwordspacing
Y.~Wang, H.~Ivison, P.~Dasigi, J.~Hessel, T.~Khot, K.~Chandu, D.~Wadden, K.~MacMillan, N.~A. Smith, I.~Beltagy, and H.~Hajishirzi, ``How far can camels go? exploring the state of instruction tuning on open resources,'' in \emph{Thirty-seventh Conference on Neural Information Processing Systems Datasets and Benchmarks Track}, 2023. [Online]. Available: \url{https://openreview.net/forum?id=w4zZNC4ZaV}
\BIBentrySTDinterwordspacing

\bibitem{wang-etal-2024-answer}
\BIBentryALTinterwordspacing
Y.~Wang, H.~Li, X.~Han, P.~Nakov, and T.~Baldwin, ``Do-not-answer: Evaluating safeguards in {LLM}s,'' in \emph{Findings of the Association for Computational Linguistics: EACL 2024}, Y.~Graham and M.~Purver, Eds.\hskip 1em plus 0.5em minus 0.4em\relax St. Julian{'}s, Malta: Association for Computational Linguistics, Mar. 2024, pp. 896--911. [Online]. Available: \url{https://aclanthology.org/2024.findings-eacl.61}
\BIBentrySTDinterwordspacing

\bibitem{touvron2023llama}
\BIBentryALTinterwordspacing
H.~Touvron, T.~Lavril, G.~Izacard, X.~Martinet, M.-A. Lachaux, T.~Lacroix, B.~Rozi{\`e}re, N.~Goyal, E.~Hambro, F.~Azhar \emph{et~al.}, ``Llama: Open and efficient foundation language models,'' \emph{ArXiv preprint}, vol. abs/2302.13971, 2023. [Online]. Available: \url{https://arxiv.org/abs/2302.13971}
\BIBentrySTDinterwordspacing

\bibitem{nvidia_megatron}
\BIBentryALTinterwordspacing
D.~Narayanan, M.~Shoeybi, J.~Casper, P.~LeGresley, M.~Patwary, V.~Korthikanti, D.~Vainbrand, P.~Kashinkunti, J.~Bernauer, B.~Catanzaro, A.~Phanishayee, and M.~Zaharia, ``Efficient large-scale language model training on gpu clusters using megatron-lm,'' in \emph{Proceedings of the International Conference for High Performance Computing, Networking, Storage and Analysis}, ser. SC '21.\hskip 1em plus 0.5em minus 0.4em\relax New York, NY, USA: Association for Computing Machinery, 2021. [Online]. Available: \url{https://doi.org/10.1145/3458817.3476209}
\BIBentrySTDinterwordspacing

\bibitem{epfmgtrn}
\BIBentryALTinterwordspacing
A.~H. Cano, M.~Pagliardini, A.~Köpf, K.~Matoba, A.~Mohtashami, O.~S. Fan, A.~Marmet, D.~Bayazit, I.~Krawczuk, Z.~Chen, F.~Salvi, A.~Bosselut, and M.~Jaggi, ``epfllm megatron-lm,'' 2023. [Online]. Available: \url{https://github.com/epfLLM/Megatron-LLM}
\BIBentrySTDinterwordspacing

\bibitem{chen2023meditron70b}
\BIBentryALTinterwordspacing
Z.~Chen, A.~H. Cano, A.~Romanou, A.~Bonnet, K.~Matoba, F.~Salvi, M.~Pagliardini, S.~Fan, A.~Köpf, A.~Mohtashami, A.~Sallinen, A.~Sakhaeirad, V.~Swamy, I.~Krawczuk, D.~Bayazit, A.~Marmet, S.~Montariol, M.-A. Hartley, M.~Jaggi, and A.~Bosselut, ``Meditron-70b: Scaling medical pretraining for large language models,'' 2023. [Online]. Available: \url{https://arxiv.org/abs/2311.16079}
\BIBentrySTDinterwordspacing

\bibitem{zhou2023lima}
\BIBentryALTinterwordspacing
C.~Zhou, P.~Liu, P.~Xu, S.~Iyer, J.~Sun, Y.~Mao, X.~Ma, A.~Efrat, P.~Yu, L.~YU, S.~Zhang, G.~Ghosh, M.~Lewis, L.~Zettlemoyer, and O.~Levy, ``{LIMA}: Less is more for alignment,'' 2023. [Online]. Available: \url{https://openreview.net/forum?id=KBMOKmX2he}
\BIBentrySTDinterwordspacing

\bibitem{kim2024prometheus}
\BIBentryALTinterwordspacing
S.~Kim, J.~Suk, S.~Longpre, B.~Y. Lin, J.~Shin, S.~Welleck, G.~Neubig, M.~Lee, K.~Lee, and M.~Seo, ``Prometheus 2: An open source language model specialized in evaluating other language models,'' 2024. [Online]. Available: \url{https://arxiv.org/abs/2405.01535}
\BIBentrySTDinterwordspacing

\bibitem{song-etal-2024-finesure}
\BIBentryALTinterwordspacing
H.~Song, H.~Su, I.~Shalyminov, J.~Cai, and S.~Mansour, ``{F}ine{S}ur{E}: Fine-grained summarization evaluation using {LLM}s,'' in \emph{Proceedings of the 62nd Annual Meeting of the Association for Computational Linguistics (Volume 1: Long Papers)}, L.-W. Ku, A.~Martins, and V.~Srikumar, Eds.\hskip 1em plus 0.5em minus 0.4em\relax Bangkok, Thailand: Association for Computational Linguistics, Aug. 2024, pp. 906--922. [Online]. Available: \url{https://aclanthology.org/2024.acl-long.51}
\BIBentrySTDinterwordspacing

\bibitem{tamura-etal-2011-extractive}
\BIBentryALTinterwordspacing
A.~Tamura, K.~Ishikawa, M.~Saikou, and M.~Tsuchida, ``Extractive summarization method for contact center dialogues based on call logs,'' in \emph{Proceedings of 5th International Joint Conference on Natural Language Processing}, H.~Wang and D.~Yarowsky, Eds.\hskip 1em plus 0.5em minus 0.4em\relax Chiang Mai, Thailand: Asian Federation of Natural Language Processing, Nov. 2011, pp. 500--508. [Online]. Available: \url{https://aclanthology.org/I11-1056}
\BIBentrySTDinterwordspacing

\bibitem{favre-etal-2015-call}
\BIBentryALTinterwordspacing
B.~Favre, E.~Stepanov, J.~Trione, F.~B{\'e}chet, and G.~Riccardi, ``Call centre conversation summarization: A pilot task at multiling 2015,'' in \emph{Proceedings of the 16th Annual Meeting of the Special Interest Group on Discourse and Dialogue}, A.~Koller, G.~Skantze, F.~Jurcicek, M.~Araki, and C.~P. Rose, Eds.\hskip 1em plus 0.5em minus 0.4em\relax Prague, Czech Republic: Association for Computational Linguistics, Sep. 2015, pp. 232--236. [Online]. Available: \url{https://aclanthology.org/W15-4633}
\BIBentrySTDinterwordspacing

\bibitem{goyal2022news}
\BIBentryALTinterwordspacing
T.~Goyal, J.~J. Li, and G.~Durrett, ``News summarization and evaluation in the era of gpt-3,'' 2022. [Online]. Available: \url{https://arxiv.org/abs/2209.12356}
\BIBentrySTDinterwordspacing

\bibitem{zhang-etal-2024-benchmarking}
\BIBentryALTinterwordspacing
T.~Zhang, F.~Ladhak, E.~Durmus, P.~Liang, K.~McKeown, and T.~B. Hashimoto, ``Benchmarking large language models for news summarization,'' \emph{Transactions of the Association for Computational Linguistics}, vol.~12, pp. 39--57, 2024. [Online]. Available: \url{https://aclanthology.org/2024.tacl-1.3}
\BIBentrySTDinterwordspacing

\bibitem{ramprasad-etal-2024-analyzing}
\BIBentryALTinterwordspacing
S.~Ramprasad, E.~Ferracane, and Z.~Lipton, ``Analyzing {LLM} behavior in dialogue summarization: Unveiling circumstantial hallucination trends,'' in \emph{Proceedings of the 62nd Annual Meeting of the Association for Computational Linguistics (Volume 1: Long Papers)}, L.-W. Ku, A.~Martins, and V.~Srikumar, Eds.\hskip 1em plus 0.5em minus 0.4em\relax Bangkok, Thailand: Association for Computational Linguistics, Aug. 2024, pp. 12\,549--12\,561. [Online]. Available: \url{https://aclanthology.org/2024.acl-long.677}
\BIBentrySTDinterwordspacing

\bibitem{chen-etal-2021-dialogsum}
\BIBentryALTinterwordspacing
Y.~Chen, Y.~Liu, L.~Chen, and Y.~Zhang, ``{D}ialog{S}um: {A} real-life scenario dialogue summarization dataset,'' in \emph{Findings of the Association for Computational Linguistics: ACL-IJCNLP 2021}, C.~Zong, F.~Xia, W.~Li, and R.~Navigli, Eds.\hskip 1em plus 0.5em minus 0.4em\relax Online: Association for Computational Linguistics, Aug. 2021, pp. 5062--5074. [Online]. Available: \url{https://aclanthology.org/2021.findings-acl.449}
\BIBentrySTDinterwordspacing

\bibitem{asi-etal-2022-end}
\BIBentryALTinterwordspacing
A.~Asi, S.~Wang, R.~Eisenstadt, D.~Geckt, Y.~Kuper, Y.~Mao, and R.~Ronen, ``An end-to-end dialogue summarization system for sales calls,'' in \emph{Proceedings of the 2022 Conference of the North American Chapter of the Association for Computational Linguistics: Human Language Technologies: Industry Track}, A.~Loukina, R.~Gangadharaiah, and B.~Min, Eds.\hskip 1em plus 0.5em minus 0.4em\relax Hybrid: Seattle, Washington + Online: Association for Computational Linguistics, Jul. 2022, pp. 45--53. [Online]. Available: \url{https://aclanthology.org/2022.naacl-industry.6}
\BIBentrySTDinterwordspacing

\bibitem{mullick-etal-2024-persona}
\BIBentryALTinterwordspacing
A.~Mullick, S.~Bose, R.~Saha, A.~Bhowmick, P.~Goyal, N.~Ganguly, P.~Dey, and R.~Kokku, ``On the persona-based summarization of domain-specific documents,'' in \emph{Findings of the Association for Computational Linguistics ACL 2024}, L.-W. Ku, A.~Martins, and V.~Srikumar, Eds.\hskip 1em plus 0.5em minus 0.4em\relax Bangkok, Thailand and virtual meeting: Association for Computational Linguistics, Aug. 2024, pp. 14\,291--14\,307. [Online]. Available: \url{https://aclanthology.org/2024.findings-acl.849}
\BIBentrySTDinterwordspacing

\bibitem{fan-etal-2018-controllable}
\BIBentryALTinterwordspacing
A.~Fan, D.~Grangier, and M.~Auli, ``Controllable abstractive summarization,'' in \emph{Proceedings of the 2nd Workshop on Neural Machine Translation and Generation}, A.~Birch, A.~Finch, T.~Luong, G.~Neubig, and Y.~Oda, Eds.\hskip 1em plus 0.5em minus 0.4em\relax Melbourne, Australia: Association for Computational Linguistics, Jul. 2018, pp. 45--54. [Online]. Available: \url{https://aclanthology.org/W18-2706}
\BIBentrySTDinterwordspacing

\bibitem{jie-etal-2024-prompt}
\BIBentryALTinterwordspacing
R.~Jie, X.~Meng, L.~Shang, X.~Jiang, and Q.~Liu, ``Prompt-based length controlled generation with multiple control types,'' in \emph{Findings of the Association for Computational Linguistics ACL 2024}, L.-W. Ku, A.~Martins, and V.~Srikumar, Eds.\hskip 1em plus 0.5em minus 0.4em\relax Bangkok, Thailand and virtual meeting: Association for Computational Linguistics, Aug. 2024, pp. 1067--1085. [Online]. Available: \url{https://aclanthology.org/2024.findings-acl.63}
\BIBentrySTDinterwordspacing

\bibitem{yuan2024followinglengthconstraintsinstructions}
\BIBentryALTinterwordspacing
W.~Yuan, I.~Kulikov, P.~Yu, K.~Cho, S.~Sukhbaatar, J.~Weston, and J.~Xu, ``Following length constraints in instructions,'' 2024. [Online]. Available: \url{https://arxiv.org/abs/2406.17744}
\BIBentrySTDinterwordspacing

\bibitem{Wu2023bloombergGPT}
\BIBentryALTinterwordspacing
S.~Wu, O.~Irsoy, S.~Lu, V.~Dabravolski, M.~Dredze, S.~Gehrmann, P.~Kambadur, D.~Rosenberg, and G.~Mann, ``Bloomberggpt: A large language model for finance,'' 2023. [Online]. Available: \url{https://arxiv.org/abs/2303.17564}
\BIBentrySTDinterwordspacing

\end{thebibliography}

\end{document}